\documentclass[letterpaper]{article} 
\usepackage{aaai25}  
\usepackage{times}  
\usepackage{helvet}  
\usepackage{courier}  
\usepackage[hyphens]{url}  
\usepackage{graphicx} 
\usepackage{natbib}  
\usepackage{caption} 
\usepackage{microtype}
\usepackage{soul}
\usepackage{algorithm}
\usepackage{algorithmic}
\usepackage{lipsum} 
\usepackage{amsmath}
\DeclareMathOperator*{\argmin}{arg\,min}
\usepackage{amsfonts}
\usepackage{booktabs}
\usepackage{multirow}
\usepackage{adjustbox}
\usepackage{longtable}
\usepackage{microtype}
\usepackage{pifont} 
\newcommand{\cmark}{\ding{51}} 
\usepackage{graphicx}
\usepackage{subcaption}
\usepackage{newfloat}
\usepackage{listings}
\DeclareCaptionStyle{ruled}{labelfont=normalfont,labelsep=colon,strut=off} 
\floatstyle{ruled}
\newfloat{listing}{tb}{lst}{}
\floatname{listing}{Listing}
\title{Distribution-Level Feature Distancing for Machine Unlearning:\\ Towards a Better Trade-off Between Model Utility and Forgetting}

\author {
    Dasol Choi\textsuperscript{\rm 1,\rm 2},
    Dongbin Na\textsuperscript{\rm 3}\thanks{Corresponding Author}
}
\affiliations {
    \textsuperscript{\rm 1}Yonsei University\\
    \textsuperscript{\rm 2}MODULABS\\
    \textsuperscript{\rm 3}POSTECH\\
    dasolchoi@yonsei.ac.kr, dongbinna@postech.ac.kr
}

\begin{document}

\maketitle

\begin{abstract}
With the explosive growth of deep learning applications and increasing privacy concerns, the right to be forgotten has become a critical requirement in various AI industries. For example, given a facial recognition system, some individuals may wish to remove their personal data that might have been used in the training phase. Unfortunately, deep neural networks sometimes unexpectedly leak personal identities, making this removal challenging. 
While recent machine unlearning algorithms aim to enable models to forget specific data, we identify an unintended utility drop—correlation collapse—in which the essential correlations between image features and true labels weaken during the forgetting process.
To address this challenge, we propose Distribution-Level Feature Distancing (DLFD), a novel method that efficiently forgets instances while preserving task-relevant feature correlations. Our method synthesizes data samples by optimizing the feature distribution to be distinctly different from that of forget samples, achieving effective results within a single training epoch. Through extensive experiments on facial recognition datasets, we demonstrate that our approach significantly outperforms state-of-the-art machine unlearning methods in both forgetting performance and model utility preservation.
\end{abstract}

%

\section{Introduction}

Deep neural network models have achieved remarkable success in various computer vision applications~\cite{ResNet, EfficientNet, VITSurvey, DenseNet, YoloReview}.
Especially, recent works show large-scale foundation models demonstrate superior classification performance across a range of tasks~\cite{CLIP, BiT, GPT3, VITSurvey, GPTUnderstand}.
However, alongside these advancements, concerns emerge regarding the unintentional leakage of sensitive information, such as personal identities from training data~\cite{MembershipInferenceAttack, MIASurvey}.

Machine unlearning has emerged as a promising solution to mitigate potential data leakage~\cite{UNSIR, Neggrad, Adaptive, SISA, SSD}, particularly in upholding the \textit{right to be forgotten}, which allows individuals to request the removal of their personal information from trained models. 
For example, in medical AI applications, a patient might request that their medical images, used during the training of a diagnostic model, be removed to protect their privacy. In such a scenario, machine unlearning enables the model to forget the patient’s data without compromising overall performance on other tasks.
This growing need for privacy has driven interest in machine unlearning research within various AI-driven industries.

Despite advancements in machine unlearning algorithms, we identify a critical issue that has not been fully explored: the risk of \textbf{correlation collapse}. When simply applying existing error-maximizing methods~\cite{UNSIR, SCRUB, Zeroshot}, unexpected outcomes can occur. For instance, these methods can inadvertently increase the magnitude of loss excessively, leading to additional data leakage by making certain data points appear \textit{special}. Moreover, relying solely on these approaches may degrade the generalization performance of the model on the original task, introducing a trade-off between model utility and forgetting. We believe this degradation is due to correlation collapse, where the useful correlations between image features and their true labels are weakened. To prevent these unexpected performance drops, it is crucial to carefully adapt and improve upon the existing methods.

To address this challenge, we propose a novel framework, \textit{Distribution-Level Feature Distancing (DLFD)} that enables unlearning of specific images while maintaining the accuracy of the original task. 
Our approach shifts the \textbf{feature distribution of the retain images} away from the distribution of the forget images, by leveraging the \textit{Optimal Transport (OT)} problem~\cite{ComputationalOT, RobustOT, Sinkhorn, NearLinarSinkhorn}. Specifically, DLFD generates perturbed images by maximizing the distance between the optimized data distribution and the forget data distribution in the feature space using OT loss.

Our method demonstrates superior performance compared to state-of-the-art methods in a setting that closely reflects real-world scenarios. We also introduce and analyze the concept of correlation collapse, which has not been extensively addressed in previous works, and revisit the task-agnostic instance unlearning setting. Our contributions are as follows:

\begin{itemize}
    \item We identify and address \textbf{correlation collapse}, a critical issue that can lead to a drop in model utility, and propose an effective solution to mitigate this risk.
    \item We propose a novel method, \textit{Distribution-Level Feature Distancing (DLFD)}, that generates a proxy data distribution distinct from the distribution of data to be forgotten.
    \item Through extensive experiments, we demonstrate that our method outperforms previous SOTA methods in task-agnostic machine unlearning.
\end{itemize}

\section{Related Work}

The previous machine unlearning algorithms typically rely on two main concepts: (1) model manipulation, and (2) data manipulation.
Firstly, various studies address the machine unlearning problem by directly manipulating the parameters of the model to erase specific information.
For instance, the \textit{Fisher Forgetting}~\cite{FisherForgetting} method scrubs the model by directly adding specific noises to the parameters using the inverse of the Fisher information matrix. 
Another approach, SCRUB~\cite{SCRUB}, improves forgetting performance by using a teacher model that is a clone of the original model. This method trains the unlearned model by minimizing the KL divergence between the output probability of the unlearned model ($\theta_{unlearned}$) and that of the teacher model ($\theta_{teacher}$). 
Similarly, the BadTeaching~\cite{BadTeaching} method employs three models: a competent teacher, an incompetent teacher, and a student (unlearned model $\theta_{unlearned}$). 
The student model is trained to mimic the competent teacher on the $D_{retain}$ while following the incompetent teacher on the $D_{forget}$. These methods highlight the effectiveness of teacher-student models in enhancing unlearning performance.

On the other hand, some methods focus on data manipulation. For example, UNSIR~\cite{UNSIR} generates noise that is added to the data to maximize the loss values for a specific target class that needs to be forgotten. Training on these error-maximized data points has shown good forgetting performance. Building on this, another method~\cite{Zeroshot} uses \textit{samples to be retain} to improve unlearning scores, extending the work of UNSIR. Similarly, recent works~\cite{Instance_Unlearn} use perturbing noise to increase the loss value, focusing primarily on error-maximizing synthesized images to achieve a high forgetting score.

\begin{figure}[htp]
    \centering
    \centerline{
    \includegraphics[width=0.42\textwidth]{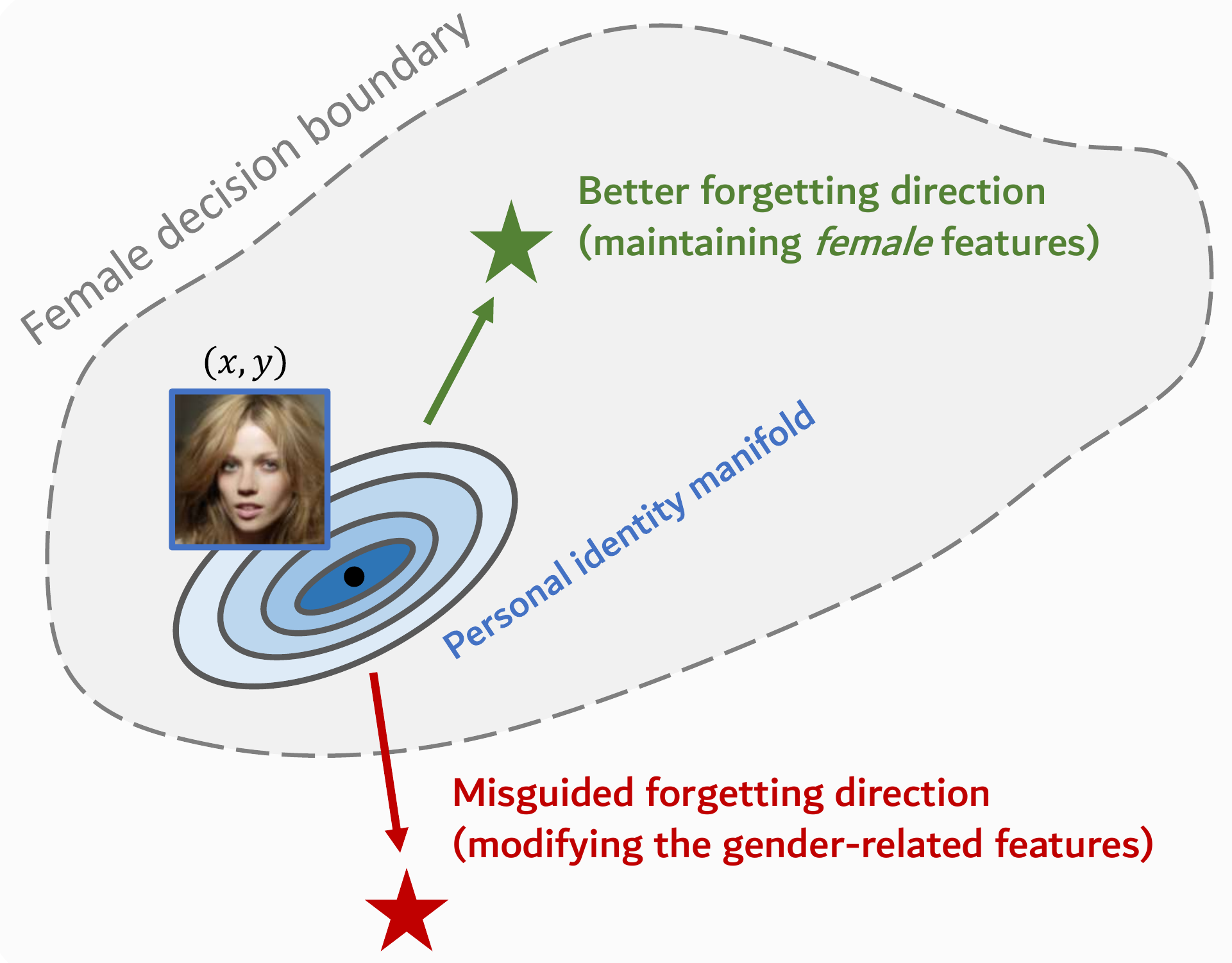}}
    \vspace{-0.5em}
    \caption{The concept of correlation collapse. If following the \textbf{misguided} forgetting direction, the correlation between the task-related useful features and labels can weaken.}
    \label{fig:collapse}
    \vspace{-1.0em}
\end{figure}

Despite their effectiveness in achieving high forgetting scores, we argue that this error-maximizing approach can easily lead to \textbf{correlation collapse} (Figure~\ref{fig:collapse}), where the useful correlations between features and labels degrade. Our work addresses these challenges by focusing on distribution-level changes rather than instance-level perturbations, which will be elaborated upon in the subsequent sections.

\section{Motivation: Correlation Collapse}

In the general computer vision domain, a feature vector \( w \in \mathcal{W} \) corresponding to an image \( x \) might contain various semantic information~\cite{LatentHSJA, encoding}. Some of these semantic features, which we denote as \( w_{\text{task}} \), are highly correlated with the original task that the model \( \theta_{\text{original}} \) is designed to solve. In addition, for personal identity unlearning tasks, another set of features, \( w_{\text{identity}} \), represents information specific to personal information. 

In the latent space \( \mathcal{W} \), we denote \( \mathcal{W}_{\text{identity}} \subset \mathcal{W} \) and \( \mathcal{W}_{\text{task}} \subset \mathcal{W} \) as the manifolds of identity and task features respectively, with \( \mathcal{W}_{\text{identity}} \cap \mathcal{W}_{\text{task}} \neq \emptyset \). This feature space overlap manifests in individual feature vectors: for any image, its feature representations \( w_{\text{identity}} \) and \( w_{\text{task}} \) share common elements. For example, in facial gender classification, attributes like hair length and facial structure exist in both identity and task-relevant features—they help identify an individual while also providing gender-related information.

This inherent overlap leads to what we term \textbf{feature entangling}, making it fundamentally challenging to separate identity information from task-relevant features. When error-maximizing methods attempt to remove identity information, they inevitably affect the shared features, resulting in \textbf{correlation collapse}: a phenomenon where the model's ability to leverage task-relevant features deteriorates, leading to degraded classification performance.

\begin{figure}[htp]
    \centering
    \begin{subfigure}[t]{0.49\columnwidth}  
        \centering
        \includegraphics[width=\textwidth]{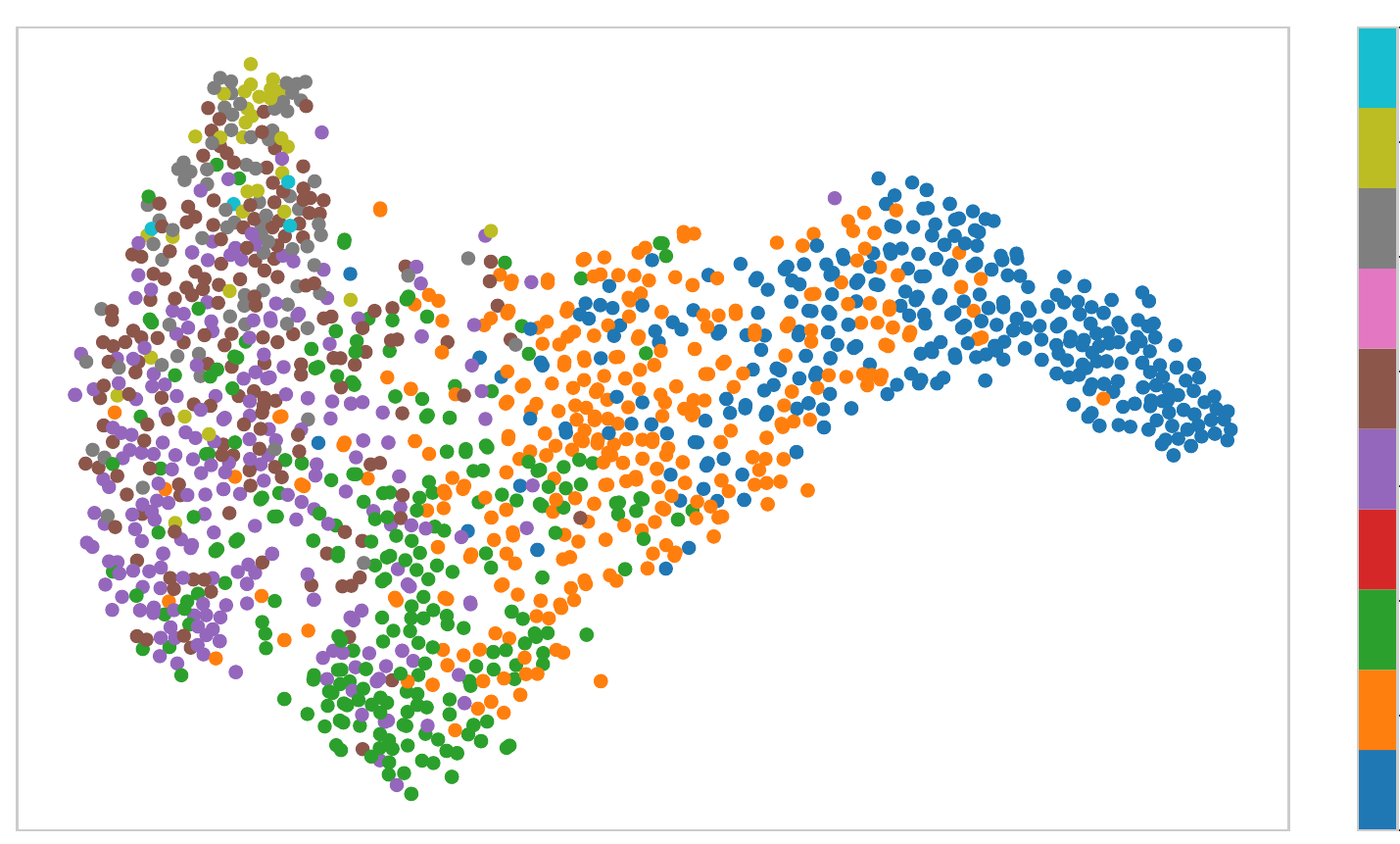}
        \caption{Original Model}
        \label{fig:original}
    \end{subfigure}
    \hfill 
    \begin{subfigure}[t]{0.49\columnwidth}  
        \centering
        \includegraphics[width=\textwidth]{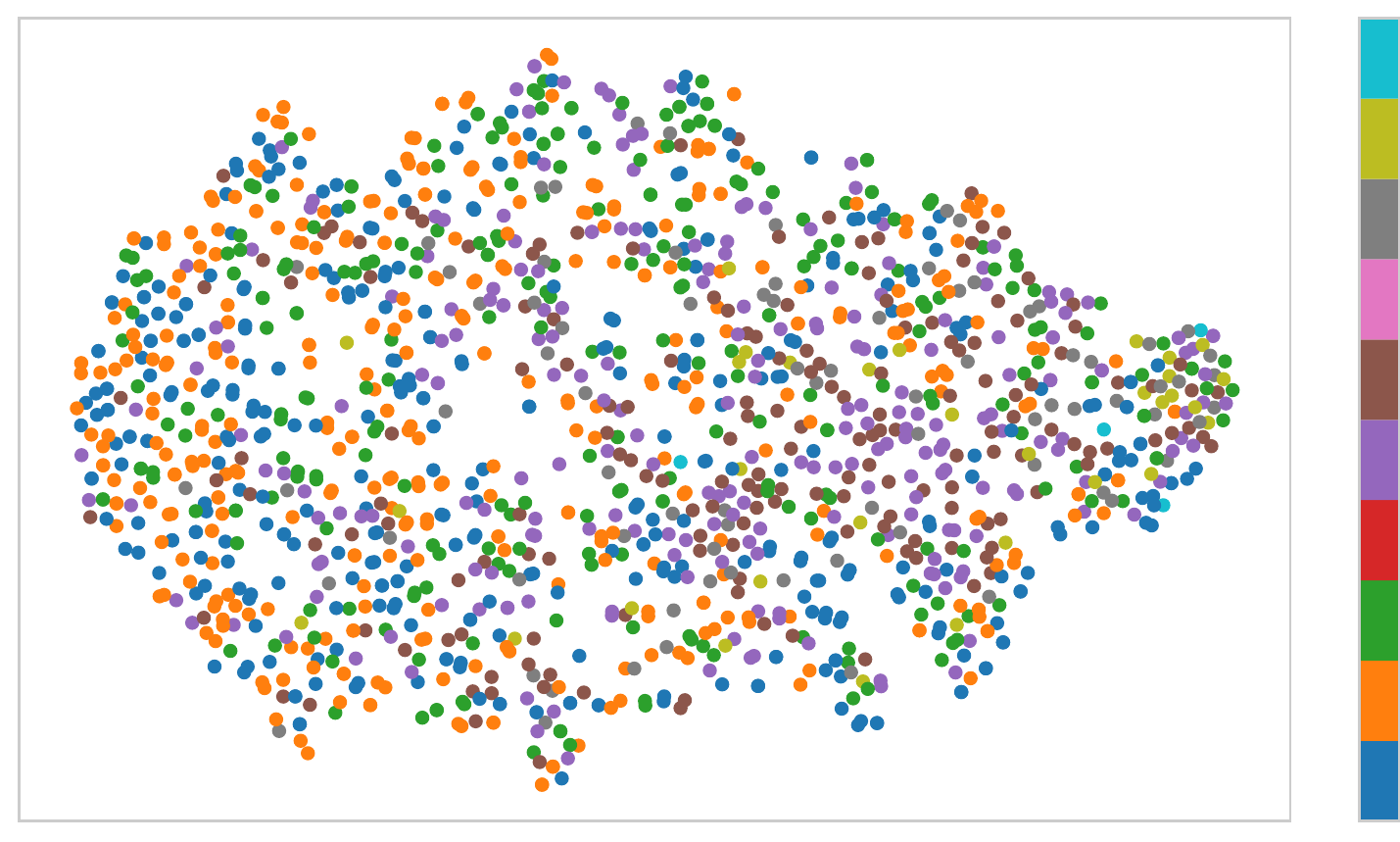}
        \caption{Error Maximized Model}
        \label{fig:error_maximizing}
    \end{subfigure}
    \vspace{-0.6em}
    \caption{Feature representations from age classification model. (a) demonstrates clear class distinctions, with age groups well-separated in feature space. (b), derived using Negative Gradient method, shows clustered features with less distinction, illustrating correlation collapse.}
    \label{fig:cls_seperated_features}
    \vspace{-0.5em}
\end{figure}

\begin{figure*}[htp]
    \centering    \centerline{\includegraphics[width=0.95\textwidth]{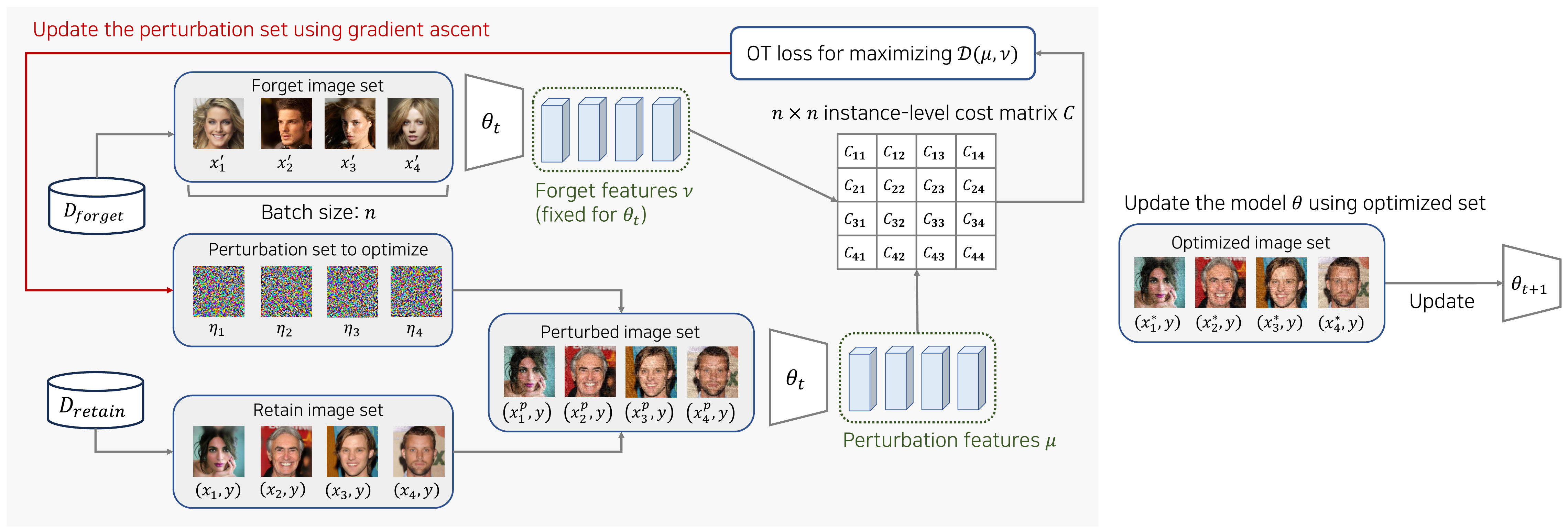}}
    \vspace{-0.6em}
    \caption{The core method of DLFD-feature distribution optimization through optimal transport. This component generates a synthesized dataset by maximizing the distance between retain and forget data distributions in the feature space. When combined with other components (detailed in Algorithm\ref{main_algorithm}), DLFD achieves a balance between model utility and forgetting performance.}
    \label{fig:main}
    \vspace{-1.0em}
\end{figure*}

As illustrated in Figure~\ref{fig:cls_seperated_features}, which visualizes feature representations in the latent space, the Original Model (Figure~\ref{fig:cls_seperated_features}(a)) maintains clear class separations, preserving task-relevant features.
In contrast, the Error Maximized model (Figure~\ref{fig:cls_seperated_features}(b)) shows diminished class distinctions, highlighting correlation collapse caused by shared feature disruption.

To address this, we propose \textbf{Distribution-Level Feature Distancing (DLFD)}, which maintains task-related features during unlearning (Figure~\ref{fig:main}). Our method preserves the structure of \( \mathcal{W}_{\text{task}} \) while modifying identity-related features.

\section{Proposed Methods}

In this section, we introduce \textbf{Distribution-Level Feature Distancing (DLFD)}, our comprehensive framework for effective machine unlearning. DLFD consists of three key components designed to balance forgetting performance and model utility.

\subsection{Feature Distribution Optimization}

Traditional approaches to machine unlearning often focus on point-wise optimization, where individual data points are manipulated to maximize the loss for data that needs to be forgotten~\cite{UNSIR, Zeroshot}. However, such methods can lead to issues like label leakage and correlation collapse, where the underlying relationships between features are disrupted~\cite{LabelLeaking, PGD, TheyAreFeatures}. To overcome these limitations, we propose a more holistic approach that considers the entire distribution of the data.

Our first component shifts the retain data distribution (\( \mu \)) away from the forget data distribution (\( \nu \)) by leveraging the optimal transport (OT) distance. Unlike simpler metrics such as KL or JS divergence, it captures the complex, high-dimensional relationships between data points~\cite{WGAN, WGANGP}.
The OT distance between the distributions \( \mu \) and \( \nu \) is defined as:

\begin{equation}
  \mathcal{D}(\mu, \nu) = \inf_{\gamma \in \prod(\mu, \nu)} \mathbb{E}_{(w, w')\sim\gamma}[c(w, w')]
\end{equation}

Here, \( \gamma \) is the set of all possible joint distributions that can transport \( \mu \) to \( \nu \), and \( c(w, w') \) represents the cost based on cosine similarity between feature vectors.
To handle the complexity of solving this problem directly, we employ a differentiable Sinkhorn method~\cite{Sinkhorn, NearLinarSinkhorn}, which approximates the solution efficiently and reduces the computational complexity to \( \mathcal{O}(n^2) \) for mini-batch computations.

To further refine the OT distance, we reformulate the problem to find an optimal transport plan \( T \):

\begin{equation}
  T^{\lambda} = \argmin_{T \in \prod(\mu, \nu)} \langle T,C \rangle - \frac{1}{\lambda} \sum_{i=1}^n \sum_{j=1}^n T_{ij} \log T_{ij}
\end{equation}

In this equation, the cost matrix \( C \) captures pairwise distances between feature vectors from \( \mu \) and \( \nu \). The regularization term \( \lambda \) keeps the transport plan \( T \) smooth, preventing overly concentrated mass transfers that could destabilize the model. Iteratively optimizing \( T^{\lambda} \) effectively separates the retain and forget data distributions, mitigating the risk of correlation collapse.

\begin{figure}[h!]
\vspace{-0.4em}
\centering
\begin{algorithm}[H]
\caption{Distribution-Level Feature Distancing (DLFD)}
\label{main_algorithm}
\begin{algorithmic}[1]
\STATE \textbf{Input:} Total batch iterations in one epoch $K$, Feature distancing steps $M$, learning rate $\gamma$, step size $\alpha$, batch size $n$, retain dataset $\mathcal{D}_{retain}$, forget dataset $\mathcal{D}_{forget}$, model $\theta_{original}$
\STATE \textbf{Output:} Unlearned model $\theta^{*}$
\STATE \textbf{Initialization:} $\theta^{*} \gets \theta_{original}$ 
\FOR{$k = 1$ to $K$} 
    \STATE \textbf{Sample retain and forget batches:} 
    \STATE \quad $\{(x_{i}, y_{i})\}_{1}^{n} \sim \mathcal{D}_{retain}$
    \STATE \quad $\{(x'_{i}, y'_{i})\}_{1}^{n} \sim \mathcal{D}_{forget}$
    \STATE \textbf{Evaluate forgetting score $F_{score}$}
    \IF{$F_{score} \geq$ threshold}
        \STATE \textbf{Initialize perturbed samples}: $\{x^{*}_{i}\}_{1}^{n} \gets \{x_{i}\}_{1}^{n}$
        \FOR{$m=1$ to $M$}
            \STATE \textbf{Compute OT loss for perturbation}: 
            \STATE \quad Extract features $F_{\text{retain}} \gets F(\{x^{*}_{i}\}_{1}^{n})$ 
            \STATE \quad $F_{\text{forget}} \gets F(\{x'_{i}\}_{1}^{n})$
            \STATE \quad $l_{OT} \gets \text{Optimal Transport loss}(F_{\text{retain}}, F_{\text{forget}})$
            \STATE \textbf{Compute classification loss}: 
            \STATE \quad $l_{CE} \gets \text{CE}(\{y_{i}\}_{1}^{n}, \theta^{*}(\{x^{*}_{i}\}_{1}^{n}))$
            \STATE \textbf{Compute combined perturbation loss}: 
            \STATE \quad $\lambda \gets \text{linear\_weight}(k, K)$
            \STATE \quad $l_{perturb} \gets l_{OT} + \lambda \cdot (- l_{CE})$

            \STATE \textbf{Update samples with perturbation loss}: 
            \STATE \quad $\{x^{*}_{i}\}_{1}^{n} \gets \{x^{*}_{i}\}_{1}^{n} + \alpha \cdot \text{sign}(\nabla_{\{x^{*}_{i}\}_{1}^{n}} l_{perturb})$
        \ENDFOR
        \STATE \textbf{Apply perturbation loss to the model}: 
        \STATE \quad $l_{train} \gets \text{CE}(\{y_{i}\}_{1}^{n}, \theta^{*}(\{x^{*}_{i}\}_{1}^{n}))$
    \ELSE
        \STATE Compute classification loss for model update: 
        \STATE \quad $l_{train} \gets \text{CE}(\{y_{i}\}_{1}^{n}, \theta^{*}(\{x_{i}\}_{1}^{n}))$
    \ENDIF
    \STATE \textbf{Update model parameters}: 
    \STATE \quad $\theta^{*} \gets \theta^{*} - \gamma \cdot \nabla_{\theta^{*}} l_{train}$
\ENDFOR
\end{algorithmic}
\end{algorithm}
\vspace{-1.7em}
\end{figure}

\subsection{Classification Loss Preservation}

To maintain model utility during the unlearning process, our second component incorporates a \textit{classification loss} guiding the perturbation process to address correlation collapse. The classification loss ensures that the original class information of the retain data is preserved, even as the model attempts to forget data points. A critical aspect of this component is the use of a \textbf{linear weight} dynamically adjusting the importance of the classification loss throughout the training process.

The linear weight plays a crucial role in balancing the trade-off between maximizing the separation of distributions and preserving the model's utility. At the beginning of training, it is set lower, allowing the model to focus more on maximizing the distance between the retain and forgot data distributions. As training progresses, the linear weight gradually increases, shifting the model's focus toward preserving the original class-specific features of the retain data.
The perturbation applied to the retain data points \( x_i \) is computed as follows:

\begin{equation}
  x_i^{*} \gets x_i + \alpha \cdot \text{sign}\left(\nabla_{x_i} \left[l_{OT} - \lambda \cdot l_{CE} \right]\right)
\end{equation}

Here, \( l_{OT} \) represents the OT loss between the retain and forget data distributions, while \( l_{CE} \) indicates the classification loss, weighted by the linear factor \( \lambda \) and computed as:

\begin{equation}
l_{CE} \gets \text{CE}(y_{i}, \theta(x^{*}_{i}))
\end{equation}

The linear weight \( \lambda \) is adjusted throughout the training process to balance the trade-off between maximizing the OT loss and preserving the classification accuracy. The perturbation is scaled by a step size \( \alpha \) and applied in the direction that increases the OT loss and decreases the weighted classification loss. This ensures that the perturbed data not only becomes more distinct from the forget data but also maintains its original task-related features.

\subsection{Dynamic Forgetting Strategy}

Our third component introduces an adaptive approach designed to optimize the forgetting process by continuously monitoring the forgetting score during training. Specifically, a subset of the validation set is used to simplify the forgetting monitoring process. When the forgetting score, assessed through this subset, drops below a predefined threshold—indicating that the model has sufficiently forgotten the target data—the algorithm dynamically shifts its focus from using the optimal transport optimization to exclusively fine-tuning the model with classification loss.

This transition not only reduces the computational overhead by avoiding unnecessary further perturbations but also ensures that the model's original task performance remains stable. By fine-tuning solely with classification loss at this stage, the strategy helps preserve the important task-related features, preventing potential degradation in model utility.

\begin{table*}[t]
\centering
\renewcommand\arraystretch{1.2}
\begin{adjustbox}{width=1\textwidth, center}
\begin{tabular}{c|c|cc||ccccccc|c} 
\toprule
\addlinespace[0.4em]
 & Evaluation Metrics & Original & Retrained & Fine-tunning &  NegGrad &  CF-k& EU-k & UNSIR~  & \begin{tabular}[c]{@{}c@{}}BadT\end{tabular} & SCRUB & \textbf{DLFD}  \\ 
\addlinespace[0.4em]
\hline
\multirow{4}{*}{\begin{minipage}{2.cm}
\centering Facial Age\\(8-classes)\end{minipage}} 
& Test Acc. $\uparrow$ & 0.6329 & 0.6050 & \textbf{0.6349} & 0.6283 & 0.6323 & 0.4767 & 0.5950 & 0.3663 & 0.6311 & 0.6166 \\
& Top-2 Acc. $\uparrow$ & 0.8803 & 0.8430 & \textbf{0.8876} & 0.8736 & 0.8736 & 0.6901 & 0.8503 & 0.6828 & 0.8743 & 0.8806 \\
& Forgetting Score $\downarrow$ & 0.1923 & 0.0767 & 0.1980 & 0.1880 & 0.1853 & 0.0438 & 0.0887 & 0.0455 & 0.1614 & \textbf{0.0385} \\
\cline{2-12}
& \textbf{NoMUS $\uparrow$} & 0.6241 & 0.7258 & 0.61945 & 0.62615 & 0.6308 & 0.69455 & 0.7088 & 0.6376 & 0.6541 & \textbf{0.7698} \\ 
\hline
\multirow{3}{*}{\begin{minipage}{2.0cm}
\centering Facial\\Emotion\\(7-classes)\end{minipage}} 
& Test Acc. $\uparrow$ & 0.7535 & 0.6897 & 0.7509 & 0.7506 & \textbf{0.7513} & 0.7511 & 0.5788 & 0.5176 & 0.7509 & 0.6613 \\
& Forgetting Score $\downarrow$ & 0.1852 & 0.0195 & 0.1735 & 0.1862 & 0.1845 & 0.1585 &\textbf{0.0192} & 0.0250 & 0.1391 & 0.0372 \\
\cline{2-12}                       
& \textbf{NoMUS $\uparrow$} & 0.6915 & 0.8253 & 0.7019 & 0.6891 & 0.6911 & 0.7171 & 0.7702 & 0.7338 & 0.73635 & \textbf{0.7934} \\    
\hline  
\multirow{3}{*}{\begin{minipage}{2.0cm}
\centering Multi-Attributes\\(3-labels)\end{minipage}}& Average Test Acc. $\uparrow$ &0.9212	&0.8700	&0.9218	&0.4487	&0.9192	&0.9189	&\textbf{0.9233}	&0.8129	&0.7057	&0.9129\\
                        & Forgetting Score $\downarrow$ 
 &0.0501	&0.0044	&0.0443	&\textbf{0.0009}	&0.04663	&0.0399	&0.0511	&0.0164	&0.0184	&0.0281\\
                        \cline{2-12}                       
                        &\textbf{NoMUS $\uparrow$} &0.9105	&0.9306	&0.9166	&0.7234	&0.9129	&0.9195	&0.9105	&0.8900	&0.8344	&\textbf{0.9283}\\    
\hline  
\multirow{3}{*}{\begin{minipage}{2.0cm}				
\centering Facial Gender\\(binary-class)\end{minipage}}& Test Acc.
$\uparrow$ & 0.9016 & 0.8493 & 0.9215 & 0.1733 & 0.9196 & \textbf{0.9216} & 0.9142 & 0.9046 & 0.9214 & 0.8997\\
                        & Forgetting Score $\downarrow$ & 0.0461 & 0.0149 & 0.0488 & 0.0895 & 0.0581 & 0.0576 & 0.0663 & 0.0453 & 0.0615 &  \textbf{0.0306}\\
                        \cline{2-12}        
                        &\textbf{NoMUS $\uparrow$} & 0.9047& 0.9097 & 0.9119 & 0.4971 & 0.9017 & 0.9031 & 0.8908 & 0.9070 & 0.8992 & \textbf{0.9192}\\   
\bottomrule
\end{tabular}
\end{adjustbox}
\vspace{-0.7em}
\caption{Overall performance of various machine unlearning methods on ResNet18 classification tasks. Our method achieves superior NoMUS scores across all tasks, with remarkable forgetting scores while maintaining competitive test accuracy. The best score is in \textbf{boldface} except for the ground-truth (\textit{Retrained}).
\\ \textit{Note}: Fine-tuning, NegGrad~\cite{Neggrad}, CF-k, EU-k~\cite{CFK}, UNSIR~\cite{UNSIR}, BadTeaching~\cite{BadTeaching}, SCRUB~\cite{SCRUB}}
\label{Tab:main_experiments}
\vspace{-1.2em}
\end{table*}

\section{Experiments}

\subsection{Preliminaries}

In machine unlearning research, an original model $\theta_{original}$ is trained on the dataset $\mathcal{D}_{train}$ to solve a specific task. To evaluate the \textbf{model utility}, we measure the classification accuracy of the model on the test set $\mathcal{D}_{test}$. If the model achieves high accuracy on $\mathcal{D}_{test}$, it is considered to have a high utility \textit{ of the model} for the original task.

The goal of an ideal machine unlearning method is to remove the images that need to be forgotten ($\mathcal{D}_{forget}$) while maintaining the original classification performance. In this study, we adopt a common machine unlearning setting where the model has access to a subset of the training data, $\mathcal{D}_{retain}$, which the AI company may still possess. Formally, we assume that the training data $\mathcal{D}_{train}$ is composed of $\mathcal{D}_{retain}$ and $\mathcal{D}_{forget}$, following the general machine unlearning setting described by \citet{Benchmark}.
Our objective is to develop a machine unlearning algorithm that makes the unlearned model $\theta_{unlearned}$ as similar as possible to the retrained model $\theta_{retrained}$, which is considered the ground truth and is trained only on $\mathcal{D}_{retain}$.

We also introduce a dataset $\mathcal{D}_{unseen}$, which is never used during the training or testing phases of the model. This dataset serves as our test set $\mathcal{D}_{test}$ and is exclusively used for evaluating the \textbf{forgetting score}. 
It is important to note that any subject targeted for unlearning should not simultaneously belong to the three datasets: $\mathcal{D}_{forget}$, $\mathcal{D}_{retain}$, and $\mathcal{D}_{unseen}$. This ensures that the subject to be forgotten is not present across multiple datasets in the machine unlearning setting.

\subsection{Task Agnostic Instance-Unlearning}

In this work, we adopt a task-agnostic machine unlearning setup, which ensures that unlearning specific target subjects does not affect the model's original functionality. 
Traditional machine unlearning research has primarily focused on class-unlearning, where entire categories (classes) are removed from the model upon a data removal request~\cite{UNSIR, Neggrad, CatastrophicallyForgetting}. While this approach works in certain scenarios, it is not applicable in all cases. For instance, in a gender classification model, removing the \textit{male} class would leave only the \textit{female} class, rendering the model ineffective for its intended purpose of gender classification. Hence, class-unlearning is not always representative of real-world needs.

To address these limitations, we propose an instance-unlearning problem setting, which targets the removal of specific personal identities or data samples without changing the overall function of the model. This approach ensures that the model's core functionality remains intact, making it more applicable to scenarios where the goal is to forget specific data without compromising the model's utility~\cite{neurips-2023-machine-unlearning, Benchmark, GLI}.

While recent studies on instance-unlearning often focus on forcing misclassification of specific instances, this deviates from a truly task-agnostic approach~\cite{liu2024model, shen2024label, Instance_Unlearn}. Our method differs by preserving the original task's functionality while ensuring that specific instances are unlearned. For instance, consider a chest X-ray (CXR) disease classification model. This model uses chest X-ray images to predict the likelihood of diseases such as tuberculosis or pneumonia. Even if all images associated with a particular patient are removed, the model should still accurately diagnose these diseases for other patients. This task-agnostic approach ensures that the model’s core functionality is preserved, making it more robust and practical for real-world applications.
Focusing on instance-unlearning within a task-agnostic framework, our method addresses a significant gap in current research, offering a solution that maintains the model's task-related performance while effectively unlearning specific instances.

\begin{table*}[h]
\centering
\renewcommand\arraystretch{1.2}
\begin{adjustbox}{width=1\textwidth,center}
\begin{tabular}{c|cccc|cccc|cccc} 
\toprule
 & \multicolumn{4}{c}{\textbf{ResNet18}} & \multicolumn{4}{c}{\textbf{DenseNet121}} & \multicolumn{4}{c}{\textbf{EfficientNetB0}} \\ 
ID & Age & Emotion & Multi-Attr. &Gender & Age & Emotion &Multi-Attr. & Gender & Age & Emotion & Multi-Attr. & Gender \\ 
\midrule
Original & 0.6242 & 0.6915 & 0.9105 & 0.9047 & 0.6813 & 0.7259 & 0.9099 & 0.9088 & 0.6419 & 0.6960 & 0.9389 & 0.8916 \\
Retrained & 0.7258 & 0.8253 & 0.9306 & 0.9097 & 0.7617 & 0.8504 & 0.9063 & 0.8901 & 0.7580 & 0.8375 & 0.9347 & 0.8964 \\
\hline
Fine-tuning & 0.6195 & 0.7019 & 0.9166 & 0.9119 & 0.6671 & 0.7260 & 0.9174 & 0.9147 & 0.6412 & 0.7026 & 0.9324 & 0.8974 \\
NegGrad & 0.6946 & 0.6891 & 0.7234 & 0.4971 & 0.7375 & 0.7289 & 0.9220 & 0.5353 & 0.6943 & 0.6938 & 0.9344 & 0.8901 \\
CF-k & 0.6262 & 0.6911 & 0.9129 & 0.9017 & 0.6677 & 0.7286 & 0.9200 & 0.9103 & 0.6478 & 0.6931 & 0.9294 & 0.8993 \\
EU-k & 0.6309 & 0.7171 & 0.9195 & 0.9031 & 0.6900 & 0.7433 & 0.9183 & 0.9101 & 0.6507 & 0.6258 & 0.9328 & 0.8908 \\
UNSIR & 0.7088 & 0.7702 & 0.9105 & 0.8908 & 0.6321 & 0.7524 & 0.9206 & 0.9017 & 0.7315 & 0.8009 & 0.9304 & 0.8876 \\
Bad Teaching& 0.6377 & 0.7338 & 0.8900 & 0.9070 & 0.6211 & 0.6736 & 0.9170 & 0.9009 & 0.7202 & 0.6711 & 0.9352 & 0.8757 \\
SCRUB & 0.6554 & 0.7363 & 0.8344 & 0.8992 & 0.6989 & 0.6727 & 0.8916 & 0.9036 & 0.6574 & 0.67245 & 0.8960 & 0.8890 \\
\hline
\textbf{DLFD} & \textbf{0.7698} & \textbf{0.7935} & \textbf{0.9283} & \textbf{0.9192} & \textbf{0.7733} & \textbf{0.7884} & \textbf{0.9433} & \textbf{0.9202} & \textbf{0.7617} & \textbf{0.7801} & \textbf{0.9529} & \textbf{0.9208} \\ 
\bottomrule
\end{tabular}
\end{adjustbox}
\vspace{-0.7em}
\caption{The overall results of the major machine unlearning methods. The results are calculated using NoMUS. Our method shows superior performance compared to SOTA methods. The best scores are in \textbf{boldface} except the ground-truth (\textit{Retrained}).}
\label{tab:architectures}
\vspace{-1.1em}
\end{table*}

\subsection{Evaluation Protocol}

In this work, we evaluate the models using two metrics: (1) model utility and (2) forgetting score. The model utility is assessed by measuring the test accuracy on $\mathcal{D}_{test}$. A high accuracy on $\mathcal{D}_{test}$ indicates that the model retains strong performance on its original task after the unlearning process.

For forgetting performance, we define a forgetting score based on the success rate of a Membership Inference Attack (MIA)~\cite{mia}. The MIA framework is formulated as follows:

\begin{equation}
    \psi(x) = \begin{cases}
        1 & \text{if } x \in \mathcal{D}_{forget} \\
        0 & \text{if } x \in \mathcal{D}_{unseen}
    \end{cases}
\end{equation}

Given $\mathcal{D}_{forget}$ and $\mathcal{D}_{unseen}$ datasets, we train a binary classifier $\psi(\cdot)$ to distinguish between them.

The classifier $\psi(\cdot)$ is trained using binary cross-entropy loss on model predictions and loss values from $\theta_{original}$. The forgetting score is then defined as:
\begin{equation}
    \text{Forgetting Score} = |\text{MIA Acc.} - 0.5| \times 2
\end{equation}
where MIA Acc. is the binary classification accuracy of $\psi(\cdot)$.

A perfect forgetting score of 0.0 indicates that the model has completely forgotten the target data, as the MIA classifier achieves only random chance (0.5) accuracy in distinguishing between forget and unseen samples.

To capture both model utility and forgetting performance in a single metric, we calculate the \textbf{Normalized Machine Unlearning Score (NoMUS)}~\cite{Benchmark} as follows:
\begin{equation}
\text{NoMUS} = \frac{1}{2} \left( P(\hat{y} = y) + (1 - \text{Forgetting Score}) \right)
\end{equation}

where $P(\hat{y} = y)$ represents the model's classification performance on $\mathcal{D}_{test}$. NoMUS ranges from 0 to 1, with higher values indicating better overall performance in both utility preservation and successful unlearning.

\subsection{Datasets}

For our experiments, we utilize three distinct facial datasets, each designed for specific classification tasks:

\begin{itemize}
    \item \textbf{Age Estimation:} The MUFAC dataset~\cite{Benchmark} contains 13,068 facial images (128$\times$128) in 8 age groups. The training set comprises 10,025 samples, with 8,525 retained and 1,500 designated for forgetting.
    \item \textbf{Emotion Recognition:} The RAF-DB dataset~\cite{RAF} contains 15,000 images across 7 emotional classes. The training set comprises 11,044 samples, with 7,730 retained and 3,314 designated for forgetting.
    \item \textbf{Multi-Attribute Classification:} The MUCAC dataset~\cite{Benchmark}, derived from CelebA~\cite{CelebA}, consists of 30,000 facial images with three binary attributes: gender, age, and expression. The training set includes 25,933 samples, with 15,385 retained and 10,548 for forgetting.
\end{itemize}

\subsection{Experimental Setup}

For experiments, we utilize various deep-neural network architectures including ResNet~\cite{ResNet}, DenseNet~\cite{DenseNet}, and EfficientNet~\cite{EfficientNet}, widely adopted in computer vision. 
To ensure fair comparison, all machine unlearning methods start from the same $\theta_{original}$ for each task.
Specifically, methods fine-tune $\theta_{unlearned}$, initialized as $\theta_{original}$, except for the Retrained model. Serving as ground truth, the Retrained model is trained from scratch on $\mathcal{D}_{retain}$, excluding data to be forgotten, to fully represent the desired unlearning outcome.

Given the computational complexity of our method, which involves calculating OT loss and performing MIA evaluations, we limit the training to a single epoch. 
Other machine unlearning methods are also trained for 1-2 epochs to ensure a fair comparison. Additionally, we find that learning rates between 0.001 and 0.005 are effective across all models and methods, consistent with previous work~\cite{UNSIR}.

\begin{figure}[htp]
    \vspace{-0.5em}
    \centering   
    \centerline{
    \includegraphics[width=0.39\textwidth]{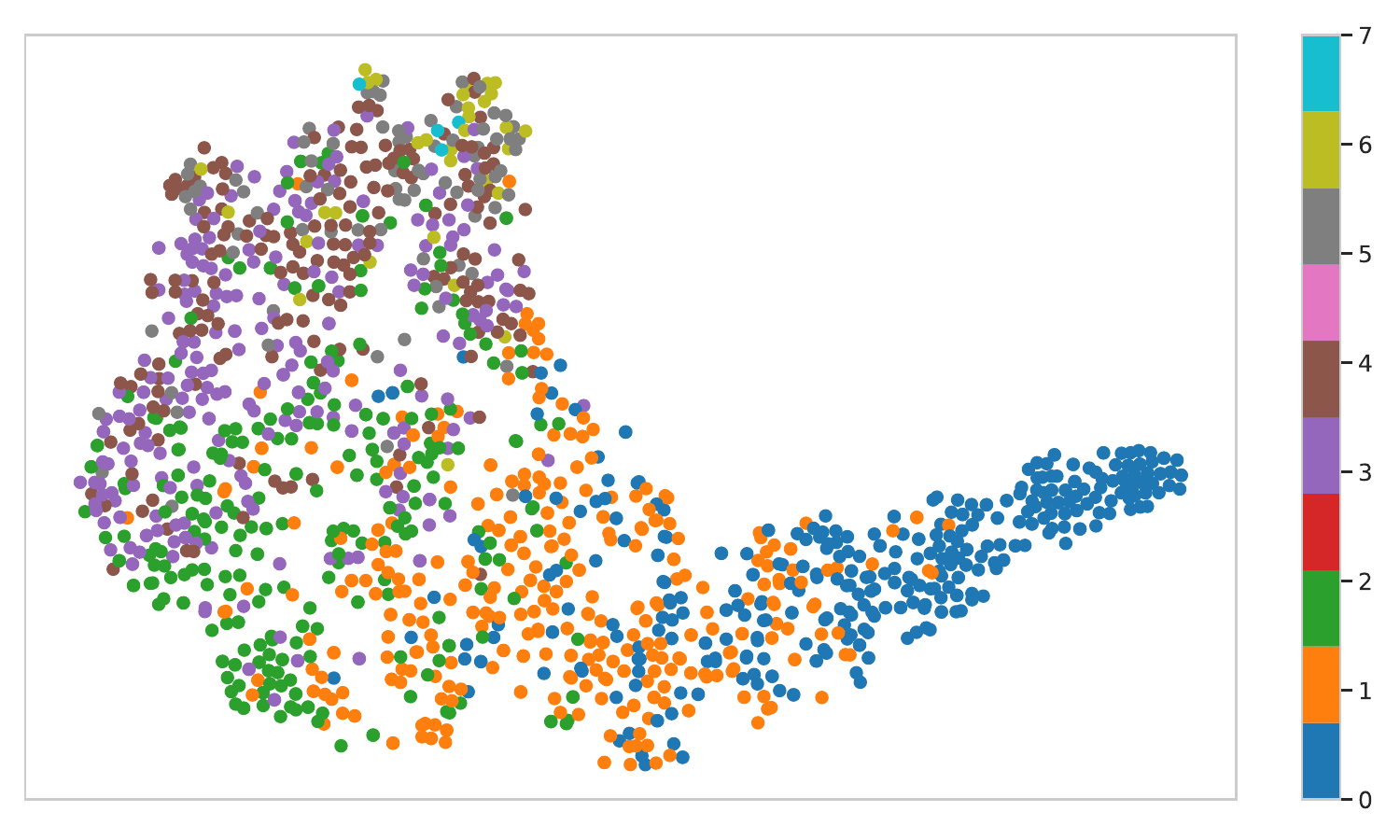}}
    \vspace{-0.8em}
    \caption{Feature representations from the age classification model trained with DLFD. The model preserves class separation similar to the original model (Figure~\ref{fig:cls_seperated_features}(a)), retaining task-relevant features while mitigating correlation collapse.}
    \label{fig:dlfd_feature}
    \vspace{-1.3em}
\end{figure}

\begin{figure*}[htp]
    \centering
    \begin{subfigure}[t]{0.33\textwidth}
        \centering
        \includegraphics[width=\textwidth]{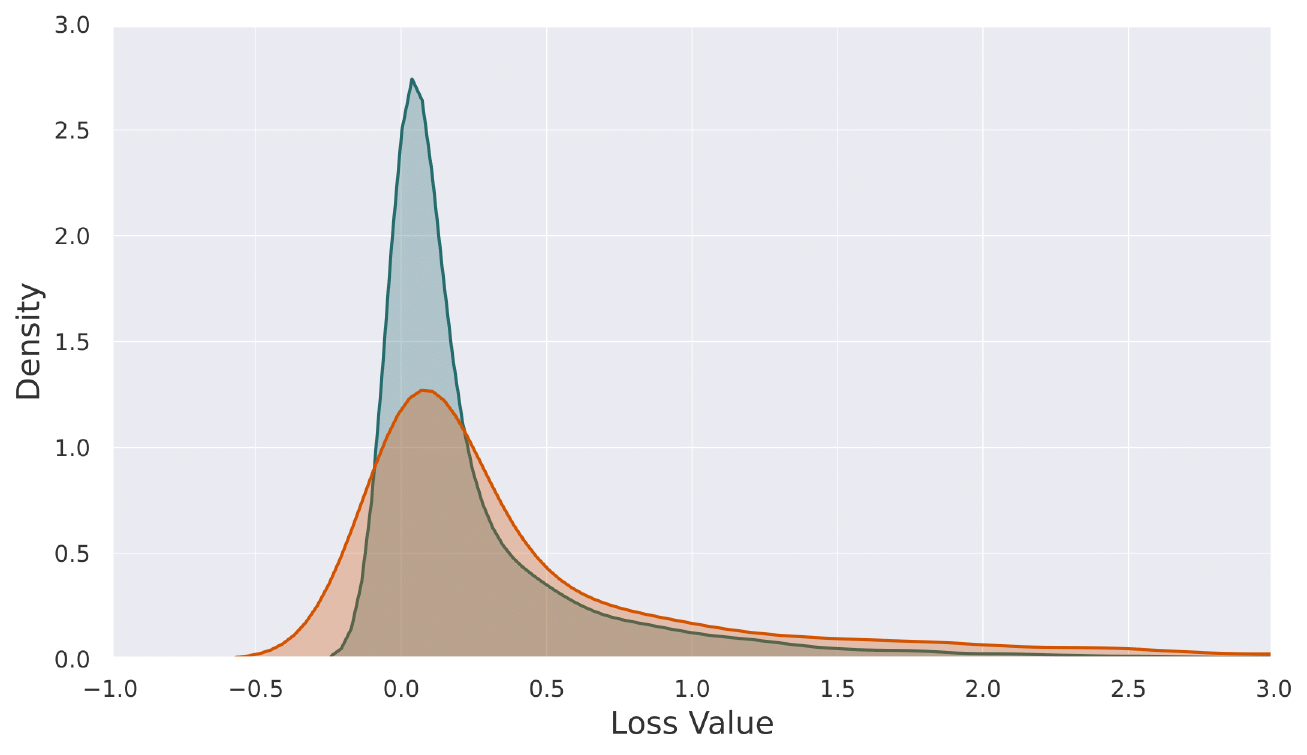}
        \vspace{-1.2em}
        \caption{\textit{Original}}
        \label{fig:original}
    \end{subfigure}
    \hfill 
    \begin{subfigure}[t]{0.33\textwidth}
        \centering
        \includegraphics[width=\textwidth]{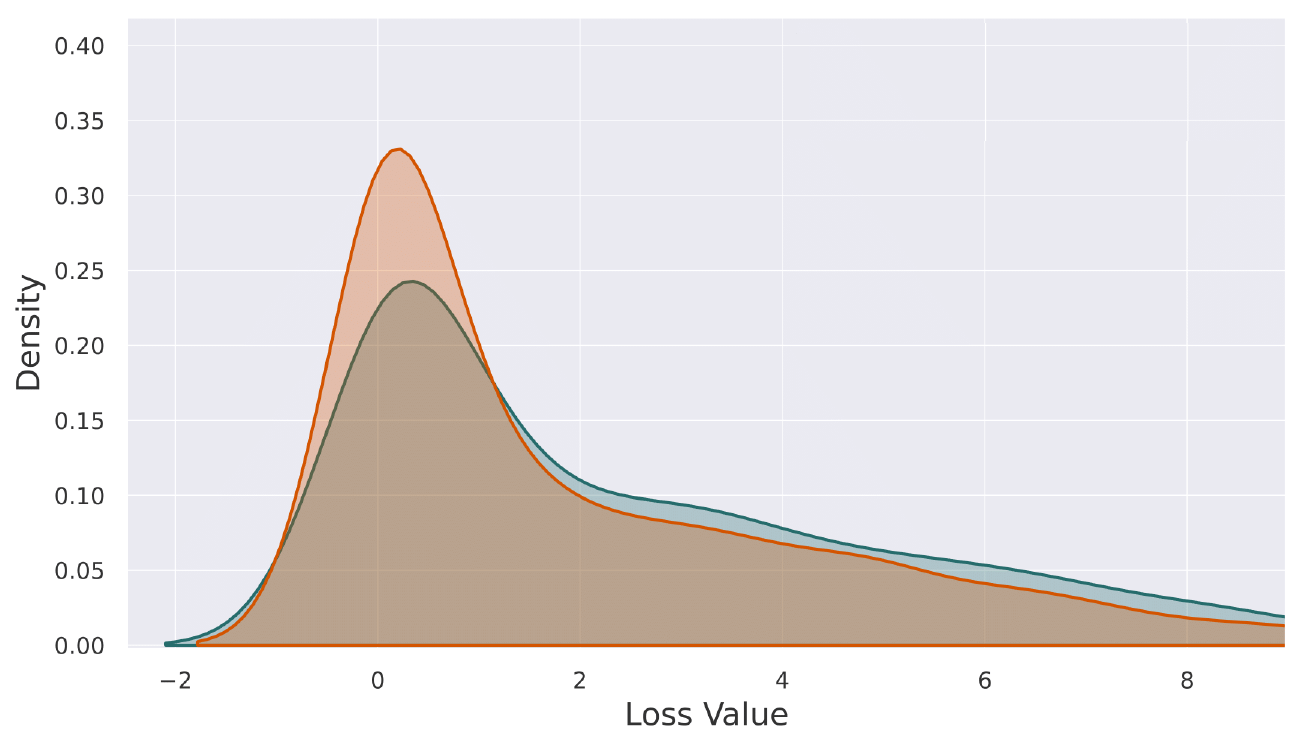}
        \vspace{-1.2em}
        \caption{\textit{Retrained}}
        \label{fig:retrained2}
    \end{subfigure}
    \hfill 
    \begin{subfigure}[t]{0.33\textwidth}
        \centering
        \includegraphics[width=\textwidth]{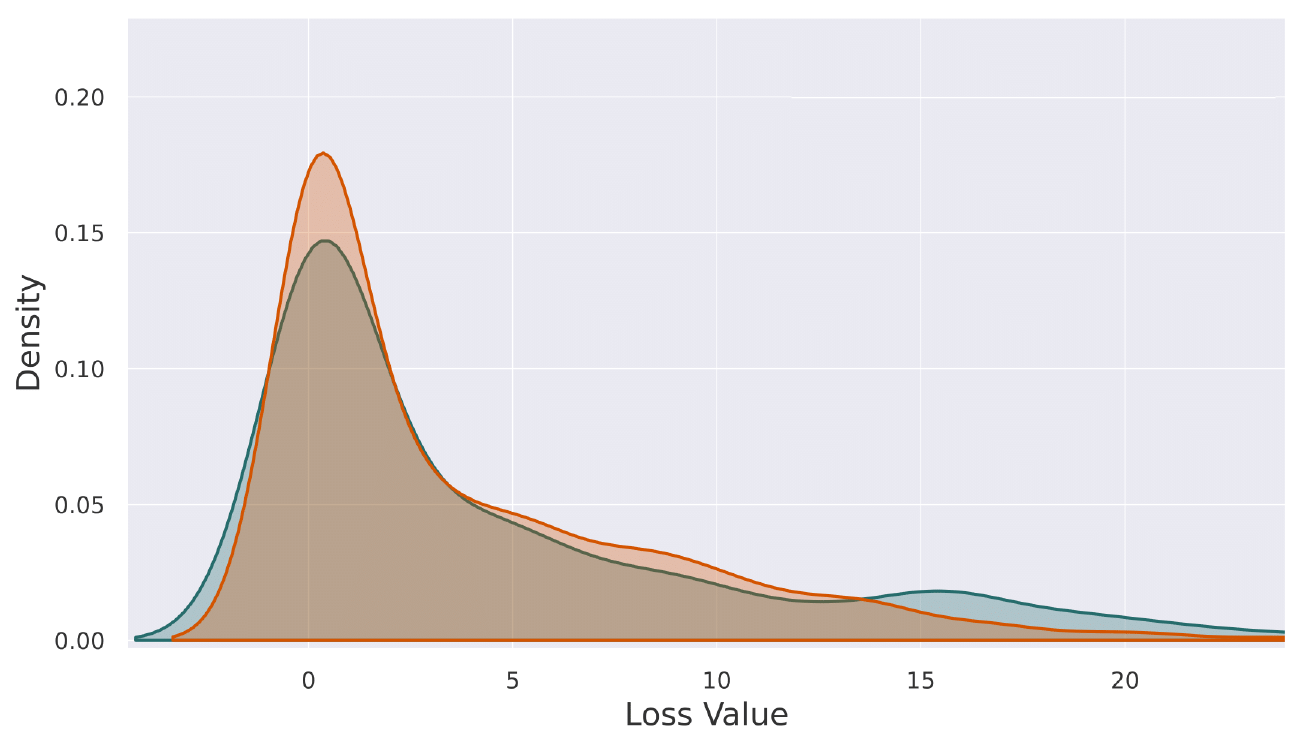}
        \vspace{-1.2em}
        \caption{\textbf{Ours}}
        \label{fig:dlfd}
    \end{subfigure}
    \vspace{-0.8em}
    \caption{The loss distributions for two baselines and ours. The orange space represents the loss distribution for unseen data, while the green represents the loss distribution for forget data. (a) illustrates loss distributions for \textit{Original} model. (b) shows loss distributions for \textit{Retrained} model. Finally (c) represents loss distributions for $\theta_{unlearned}$ fine-tuned on DLFD-optimized images.}
    \label{fig:loss_distributions}
    \vspace{-1.2em}
\end{figure*}

\subsection{Performance of DLFD Method}

We evaluate our method across four classification tasks:
facial age prediction, emotion recognition, multi-attribute classification, and gender classification. The multi-attribute model includes three binary labels: gender (female/male), age (old/young), and expression (smiling/unsmiling), with the average classification accuracy reported as the model utility. Gender classification, originally part of the multi-attribute, is also evaluated as an independent binary classification task.

We compare our method with various previously proposed methods. As shown in Table~\ref{Tab:main_experiments}, our method demonstrates superior performance in the comprehensive metric, NoMUS. Although \textit{Fine-tuning}, \textit{CF-$k$}, and \textit{EU-$k$} can generally achieve high test accuracy, their forgetting scores are generally low, indicating insufficient unlearning performance. On the other hand, the \textit{Retrained (ground-truth)} model shows excellent forgetting performance but suffers a significant drop in test accuracy, which adversely impacts model utility.

Across all experiments, our method consistently delivers competitive or superior performance in both metrics. We demonstrate that DLFD effectively unlearns the forget data while maintaining model utility.
As shown in Figure~\ref{fig:dlfd}, the loss distributions of $\mathcal{D}{unseen}$ and $\mathcal{D}{forget}$ in our method closely resemble those of the \textit{Retrained} model, considered the ground truth (Figure~\ref{fig:retrained2}). 
This similarity indicates that the unlearning algorithm works effectively.

In particular, the DLFD method shows more substantial improvements in complex, multi-class tasks such as age estimation and emotion recognition, where the feature entanglement is more significant. Conversely, the improvements in multi-attribute and gender classification tasks are relatively smaller, likely due to the binary nature of these classifications, where the complexity of feature entanglement is inherently lower. These results highlight the effectiveness of our approach in scenarios where maintaining feature integrity amid complex and overlapping feature spaces is more challenging.

Moreover, Figure~\ref{fig:dlfd_feature} displays the feature representations extracted by the DLFD model. The figure demonstrates how DLFD maintains clear class distinctions, preventing correlation collapse and preserving essential task-related features.

\subsection{Ablation Study}

We perform ablation studies to evaluate each component of DLFD. Using only feature distribution optimization initially achieves success in machine unlearning, as shown in Table~\ref{tab:ablation}. While effective in separating retain and forget data distributions, this component alone may reduce model utility without the support of other components.

The addition of classification loss preservation as the second component significantly improves performance, especially in Age and Emotion tasks (NoMUS increased by 7.3\% and 4.7\%), highlighting its role in maintaining model utility. Finally, integrating dynamic forgetting further enhances performance, with additional improvements in Age (2.\%) and Emotion (5.3\%) tasks. This component effectively prevents correlation collapse by balancing forgetting and utility preservation.
The complete framework, combining all three components, achieves superior NoMUS scores across all tasks, showing the effectiveness of their synergistic interaction.

\begin{table}[h]
\vspace{-0.2em}
\centering
\fontsize{9}{10}\selectfont
\renewcommand\arraystretch{1.2}
\begin{adjustbox}{width=8.5cm,center}
\begin{tabular}{c|c|c|c|c|c|c} 
\toprule
\multirow{2}{*}{\begin{tabular}[c]{@{}c@{}}Feature Dist.\\ Optim.\end{tabular}} & \multirow{2}{*}{\begin{tabular}[c]{@{}c@{}}Cls\\ Loss\end{tabular}} & \multirow{2}{*}{\begin{tabular}[c]{@{}c@{}}Dynamic\\ Forgetting\end{tabular}} & \textbf{Age} &\textbf{Emotion} &\textbf{Multi-Attr.} & \textbf{Gender} \\
\cline{4-7}
 &  &  & NoMUS $\uparrow$  & NoMUS $\uparrow$ & NoMUS $\uparrow$ & NoMUS $\uparrow$ \\ 
\midrule
\cmark & & & 0.7021& 0.7199 & 0.9423 & 0.8898 \\
\cmark & \cmark & &0.7536 & 0.7536 & 0.9420 & 0.9170 \\
\cmark & \cmark & \cmark & \textbf{0.7698} & \textbf{0.7934}&\textbf{0.9433} & \textbf{0.9192} \\ 
\bottomrule
\end{tabular}
\end{adjustbox}
\vspace{-0.6em}
\caption{Ablation study results for each component in DLFD, showing cumulative performance improvements.}
\label{tab:ablation}
\vspace{-1.0em}
\end{table}

\section{Discussion}

\subsection{Information Leakage in Error-Maximization}

\begin{figure}[htp]
    \vspace{-1.0em}
    \centering   \centerline{\includegraphics[width=0.47\textwidth]{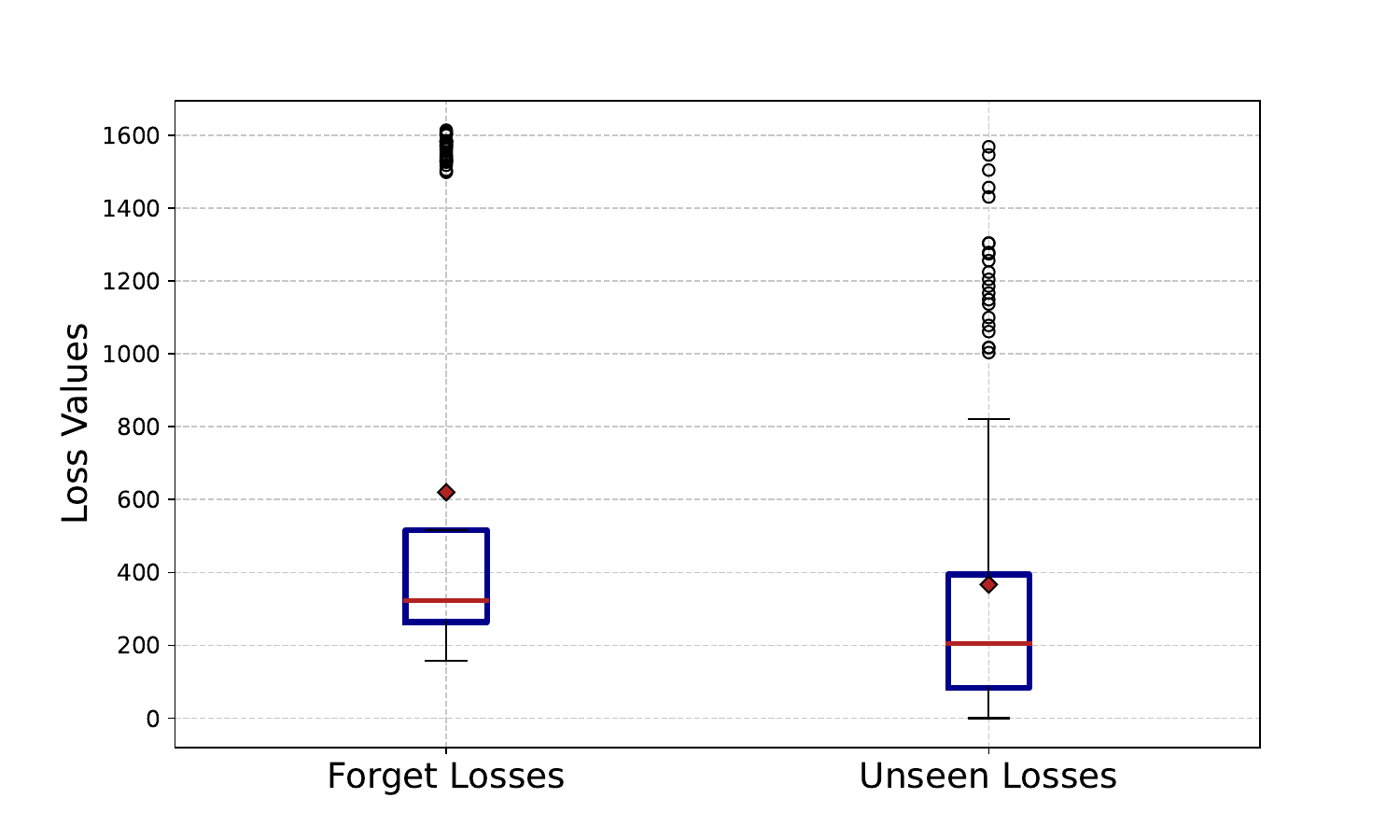}}
    \vspace{-0.9em}
    \caption{The naive instance-level error-maximizing can induce excessively high loss value for \textit{forget} data even larger than the \textit{unseen} data, which might not be desirable.}    \label{fig:forget_unseen_loss_comparision}
    \vspace{-1.2em}
\end{figure}

A trained model generally shows lower loss values for training data compared to unseen data, which can lead to data leakage. Methods like UNSIR~\cite{UNSIR} and SCRUB~\cite{SCRUB} that maximize loss for data intended to be forgotten may inadvertently increase the loss for forget data beyond that of unseen data, making the model vulnerable to membership inference attacks.
Our findings reveal that even with unlearning, naive error-maximization can still result in information leakage. Specifically, when the number of forget samples is small (\(< 100\)), the loss values for forget data can abnormally increase, exceeding those of unseen data (Figure~\ref{fig:forget_unseen_loss_comparision}). This issue highlights a risk that has been overlooked in prior studies.

\subsection{Trade-off between Model Utility and Forgetting}

Our method reveals a trade-off between test accuracy and forgetting score. As we increase the loss for the data intended to be forgotten ($x_{forget}$), the forgetting score improves, but this comes at the cost of test accuracy. This is likely due to \textit{correlation collapse}, where essential label-related features of the retain data are altered. Moreover, the effectiveness of unlearning strategies can vary depending on the dataset's characteristics, scale, and task complexity. This variability suggests the challenges of setting up robust unlearning experiments and the need for adaptive unlearning methods adjusted to different scenarios. Our findings emphasize the importance of carefully balancing model utility and forgetting performance to achieve optimal unlearning results.

\subsection{Practical Considerations and Future Work}

One potential limitation of our method could arise when retain and forget datasets have overlapping features in the feature space. While our current implementation demonstrates strong performance in settings with minimal overlap, handling heavily overlapping feature distributions remains a challenging scenario that warrants further investigation.

Moreover, although MIA is widely used as a metric to assess forgetting performance, it may not fully capture unlearning effectiveness across all scenarios. In scenarios where the model is exceptionally well-trained, the distinction between forget and unseen data may become minimal, leading to MIA scores that do not adequately reflect the true forgetting performance. This suggests the need for the unlearning community to develop more robust evaluation metrics.

\section{Conclusion}

We address key challenges in machine unlearning, including information leakage in error-maximizing methods, task-specific settings, and the critical trade-off between model utility and effective forgetting. Our proposed DLFD method effectively mitigates these issues by reducing the risk of correlation collapse while maintaining high model utility. Experimental results consistently demonstrate that DLFD outperforms existing methods across multiple benchmarks, underscoring its robustness and effectiveness.

\section{Acknowledgements}
This research was supported by Brian Impact, a non-profit organization dedicated to advancing science and technology. 

\bibliography{aaai25}

\end{document}